\documentclass[mlmain]{jmlr}

\usepackage[utf8]{inputenc}

\usepackage[hpos=300px,vpos=70px]{draftwatermark}
\SetWatermarkText{\test}
\SetWatermarkScale{1}
\SetWatermarkAngle{0}

\usepackage{longtable}

\usepackage{booktabs}
\usepackage[load-configurations=version-1]{siunitx} 

\theorembodyfont{\upshape}
\theoremheaderfont{\scshape}
\theorempostheader{:}
\theoremsep{\newline}

\jmlrvolume{}
\firstpageno{1}
\editors{List of editors' names}

\jmlryear{2025}
\jmlrworkshop{Symmetry and Geometry in Neural Representations}

\title[Representational Geometry of Color Qualia]{%
 Probing the Representational Geometry of Color Qualia: \titlebreak 
 Dissociating Pure Perception from Task Demands in Brains and AI Models%
}

 \author{
   \Name{Jing Xu} \Email{jxu101@ur.rochester.edu} \\
   \addr University of Rochester
 }

\begin{document}

\maketitle

\begin{abstract}
Probing the computational underpinnings of subjective experience, or qualia, remains a central challenge in cognitive neuroscience. This project tackles this question by performing a rigorous comparison of the representational geometry of color qualia between state-of-the-art AI models and the human brain. Using a unique fMRI dataset with a "no-report" paradigm, we use Representational Similarity Analysis (RSA) to compare diverse vision models against neural activity under two conditions: pure perception (``no-report'') and task-modulated perception (``report"). Our analysis yields three principal findings. First, nearly all models align better with neural representations of pure perception, suggesting that the cognitive processes involved in task execution are not captured by current feedforward architectures. Second, our analysis reveals a critical interaction between training paradigm and architecture, challenging the simple assumption that Contrastive Language-Image Pre-training(CLIP) training universally improves neural plausibility. In our direct comparison, this multi-modal training method enhanced brain-alignment for a vision transformer(ViT), yet had the opposite effect on a ConvNet. Our work contributes a new benchmark task for color qualia to the field, packaged in a Brain-Score compatible format. This benchmark reveals a fundamental divergence in the inductive biases of artificial and biological vision systems, offering clear guidance for developing more neurally plausible models.
\end{abstract}

\section{Introduction}
\label{sec:intro}
Deep neural networks serve as the leading computational models of the primate visual system, yet significant gaps between artificial and biological vision persist. A major research effort aims to close this gap by creating more neurally plausible models\citep{10.3389/fncom.2023.1153572}.Some approaches modify the training diet to be more biologically realistic; for instance, training models on blurry images—a common feature of natural vision—improves neural alignment, shape sensitivity, and robustness\citep{cnnrobustness}.Other approaches directly optimize models on neural data, demonstrating that ``teaching" a model to predict fMRI signals can enhance its brain-likeness \citep{CORnet}. Within this context, a foundational challenge is that the very act of reporting a percept can alter its neural representation. Seminal work by \citet{brouwer2009decoding, brouwer2013categorical} showed that the task of naming a color reshapes its neural code, a confound that complicates direct comparisons between task-optimized models and the brain's perceptual states.

To disentangle perception from the act of reporting, we use a unique fMRI dataset from \citet{hirao2025dataset} that provides neural data for color qualia under both ``no-report" (pure perception) and ``report" (task-modulated) conditions. Using Representational Similarity Analysis (RSA), we conduct a large-scale comparison between the representational geometry of over 30 AI models and these two distinct neural states. This approach allows us to contribute a new benchmark for color representation while highlighting a fundamental divergence in the inductive biases that govern artificial and biological vision.

\section{Method}
\label{sec:method}

\subsection{ Human Neuroimaging Data and Experimental Design}
We analyzed fMRI data from 20 participants in a public dataset by \citet{hirao2025dataset}. Functional images were acquired on a 3T Siemens Verio scanner with a 32-channel head coil (TR=1000ms, 3.2mm isotropic voxels). This dataset contains neural responses to nine distinct colors presented under two conditions:(1) a ``no-report''  condition, where participants passively viewed the colors, designed to isolate neural signals of pure perception; and (2) a ``report" condition, where participants performed a one-back color similarity judgment task. For each participant, we extracted the voxel patterns corresponding from the whole brain corresponding to each of the nine colors for both conditions.

\subsection{AI Vision Models}
We selected a diverse suite of over 30 state-of-the-art vision models for comparison. This suite was chosen to span a range of architectures (e.g., CNNs and Vision Transformers) and training paradigms (e.g., supervised and multi-modal CLIP). For each model, we used the Brain-Score platform\citep{brainscore}  to automatically identify the layer with the highest neural predictivity to primate area V4. For each model,we extracted feature activations from this functionally-defined layer with the highest neural predictivity in response to the same nine color stimuli.

\subsection{Representational Similarity Analysis }
To quantitatively compare the representational geometry between the AI models and the human brain, we employed RSA. The analysis, adapted from \citet{brainscore} , proceeded in two main stages: (1) construction of Representational Similarity Matrices (RSMs) for both brain and model data, and (2) quantitative comparison of these RSMs to score model-brain alignment and contextualize the results with a noise ceiling.

\begin{enumerate}
    \item \textbf{Feature Selection:} To improve the signal-to-noise ratio, we first discarded features (voxels or units) with near-zero variance. Then we selected only the most responsive features—those whose mean activation across all stimuli was greater than 3.1 standard deviations from the population mean of feature means  \citep{10.1371/journal.pcbi.1003553}.

    \item \textbf{Normalization:} The selected feature patterns were z-scored to normalize their activations before computing the RSM.

    \item \textbf{Similarity Calculation:} The final RSM was computed as the Pearson correlation matrix of these normalized patterns. Each entry $RSM_{ij}$ in the resulting $9 \times 9$ matrix represents the Pearson correlation coefficient $r$ between the brain's/model's response patterns to color $i$ and color $j$.
\end{enumerate}

To score the alignment between model and brain geometries, we vectorized the off-diagonal elements of each RSM. The model's vector was then compared against each of the 20 human subject vectors using Spearman's rank correlation. The final score is the average of these 20 correlation coefficients, a process performed separately for both the "no-report" and "report" conditions. To contextualize these scores, we estimated a 
noise ceiling using a leave-one-subject-out procedure\citep{10.1371/journal.pcbi.1003553}. This ceiling represents the theoretical upper performance bound given the data's inter-subject variability and is calculated by correlating each subject’s RSM with the average of the remaining subjects' RSMs.

\section{Result}

\label{sec:result}
By comparing the representational geometry of color qualia between AI models and human fMRI data under pure perception and task-modulated conditions, we uncovered three findings that reveal a fundamental divergence between artificial and biological vision.

\subsection{AI Models Align Better with Pure Perception }
Across the models evaluated, the average neural alignment was substantially higher for the ``no-report" condition (mean~=$0.020$) than for the ``report" condition (mean~=~$-0.010$).This general trend was highly consistent: an overwhelming majority of models—28 out of 31 (90\%)—scored higher in the ``no-report" condition.  This demonstrates that current feedforward architectures struggle to capture the geometric shifts induced by task demands.
\begin{figure}[htbp]
\floatconts
  {fig:quantitative_results}
  {\caption{\textbf{Quantitative comparison of AI models and human brain representations.} 
  \textbf{(a)} Model similarity scores to neural data under no-report (x-axis) vs. report (y-axis) conditions. The dashed line represents identity ($y=x$). Nearly all models fall below this line. 
  \textbf{(b)} Direct comparison of neural alignment for models with and without CLIP training. }}
  {%
    \subfigure[Overall Performance]{\label{fig:scatter}%
      \includegraphics[width=0.48\linewidth]{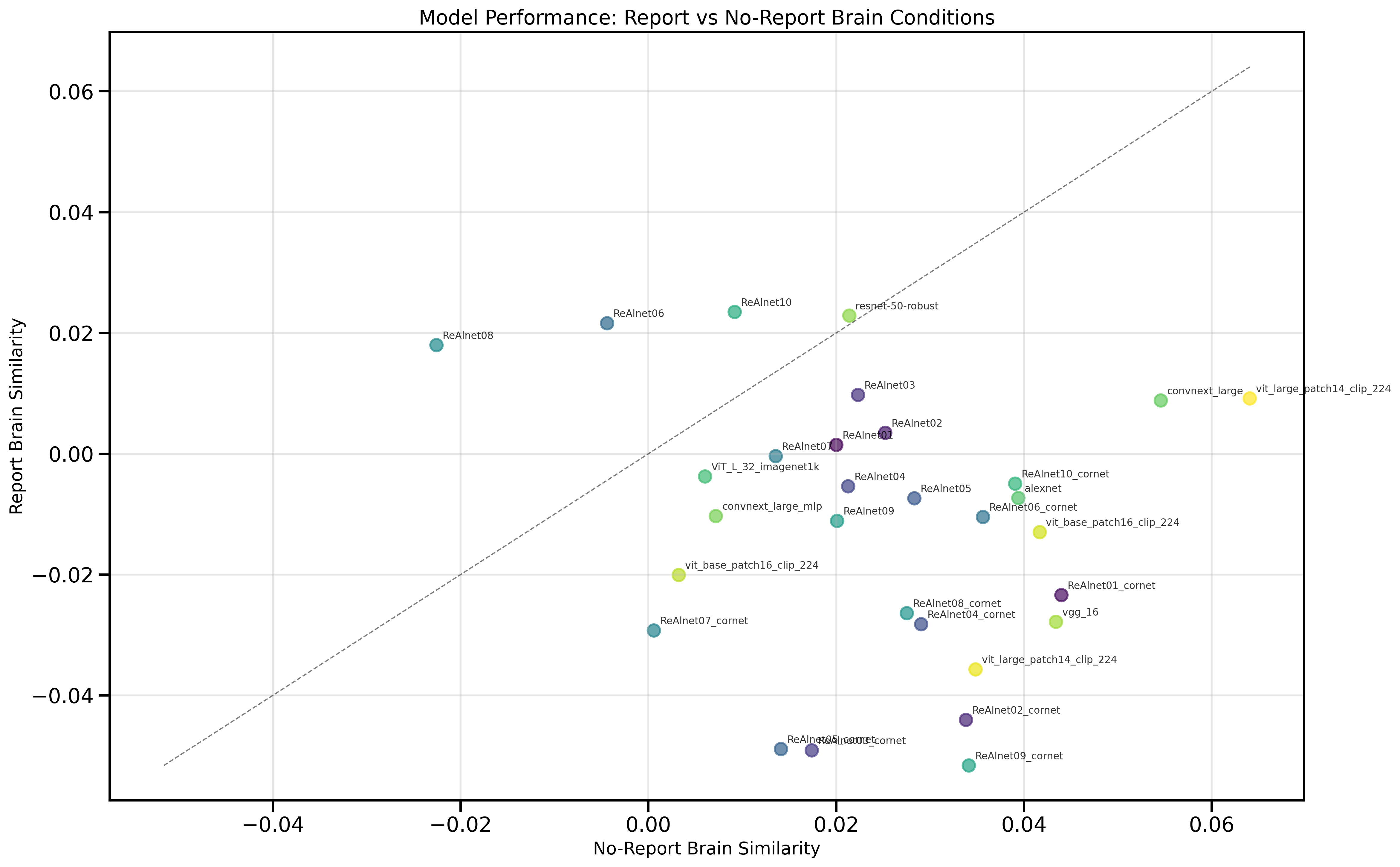}}
    \hfill
    \subfigure[CLIP vs. Supervised Training]{\label{fig:barchart}%
      \includegraphics[width=0.48\linewidth]{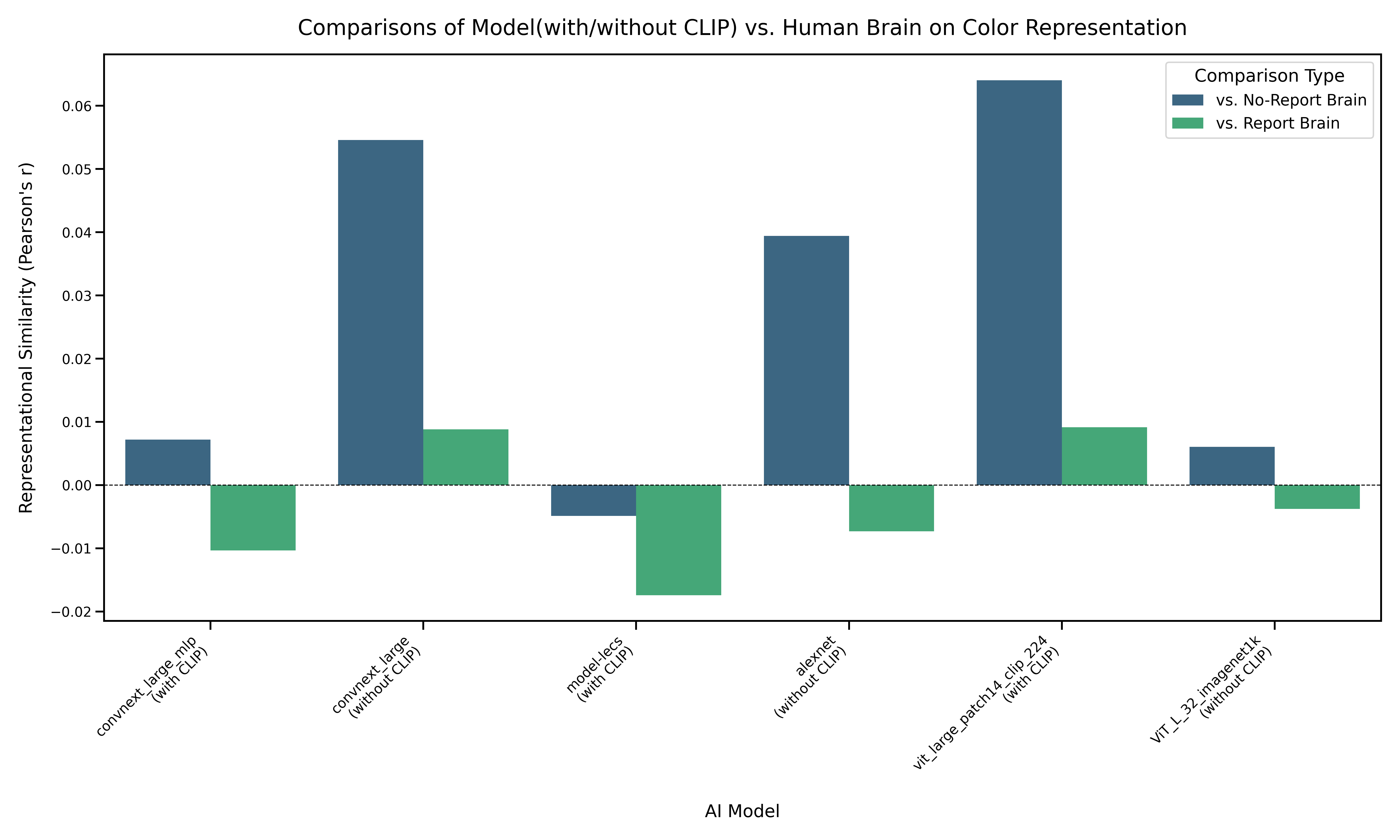}}
  }
\end{figure}

\subsection{Training Paradigms Interact with Model Architecture}
We uncover a critical interaction between training paradigm and architecture (Figure 1b). Multi-modal CLIP training enhanced neural alignment(+0.058) for a ViT but impaired it(-0.0474) for a ConvNet, relative to supervised counterparts. This challenges the assumption that CLIP training universally improves neural plausibility for basic perceptual features.

\subsection{AI Models Learn Abstract Symmetries Absent in Human Brain Representations}
We uncovered a fundamental divergence in the representational geometries learned by AI and the brain (Figure 2). High-performing AI models learn the abstract mathematical symmetry of HSV color space, evidenced by a strong anti-diagonal structure in their RDMs. This geometric pattern is absent in human neural RDMs, which are instead constrained by biological principles like color opponency. This tendency to learn abstract symmetries was general across most models we tested (see Appendix, Figure A1), revealing a profound difference in the inductive biases of artificial and biological vision.
\begin{figure}[htbp]
\floatconts
  {fig:rdm_comparison}
  {\caption{\textbf{Qualitative divergence in representational geometry.} RDMs for human brain activity and AI models. The small color suqares means what color is compared.
  The two leftmost panels show the average RDM from human fMRI data in the no-report and report conditions, respectively. The green array, with no dark purple going through, reveals they lack any clear off-diagonal structure. 
  In contrast, the RDM of the best-performing AI model (third panel) exhibits an anti-diagonal structure(blue array), reflecting the abstract mathematical symmetry of the HSV color space.}}
  {\includegraphics[width=\linewidth]{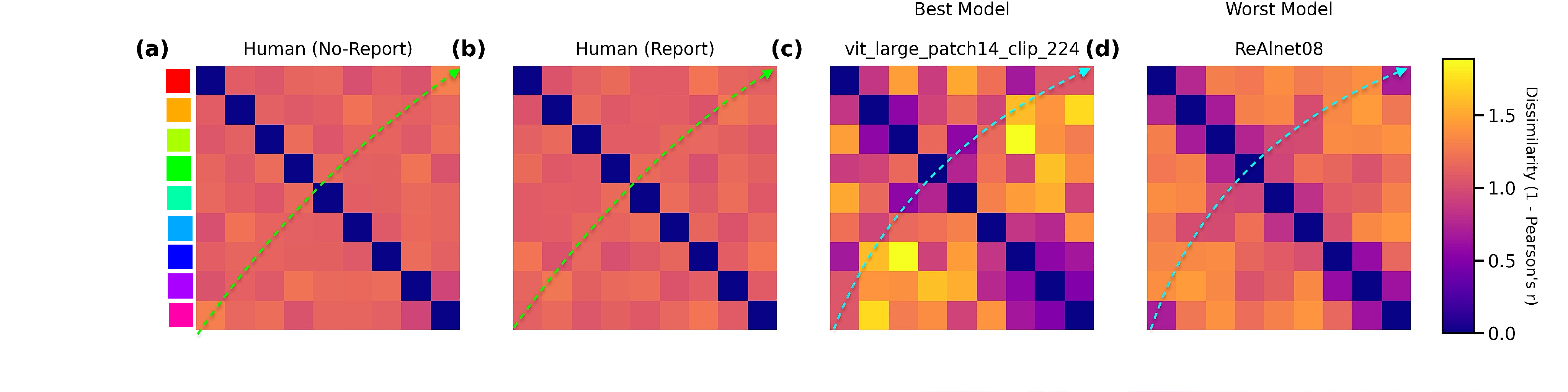}}
\end{figure}

\section{Discussion}
\label{sec:discussion}
 Future work will extend this analysis hierarchically across the visual pathway (e.g., V1, V4, IT) to map where geometric differences emerge. Additionally, a theoretical ‘HSV angle model,’ with an RSM defined purely by color angle, should be tested against our benchmark. High performance from such a simple model would confirm that current networks may rely on abstract mathematical shortcuts absent in biological vision.

\bibliography{pmlr-sample}

\appendix

\section{First Appendix}\label{apd:first}

\begin{figure}[htbp]
\floatconts
  {fig:all_rdms_appendix}
  {\caption{\textbf{RDMs for all tested models.} 
  This figure shows the $9 \times 9$ RDM for each of the AI models evaluated in this study. A prominent anti-diagonal structure, highlighted by the cyan curves on select plots, is visible in many of the high-performing models, particularly those based on ViT and CLIP training paradigms. This provides visual evidence that the tendency to learn the abstract mathematical symmetry of the stimulus set is a general phenomenon across a wide range of modern vision models.}}
  {\includegraphics[width=\linewidth]{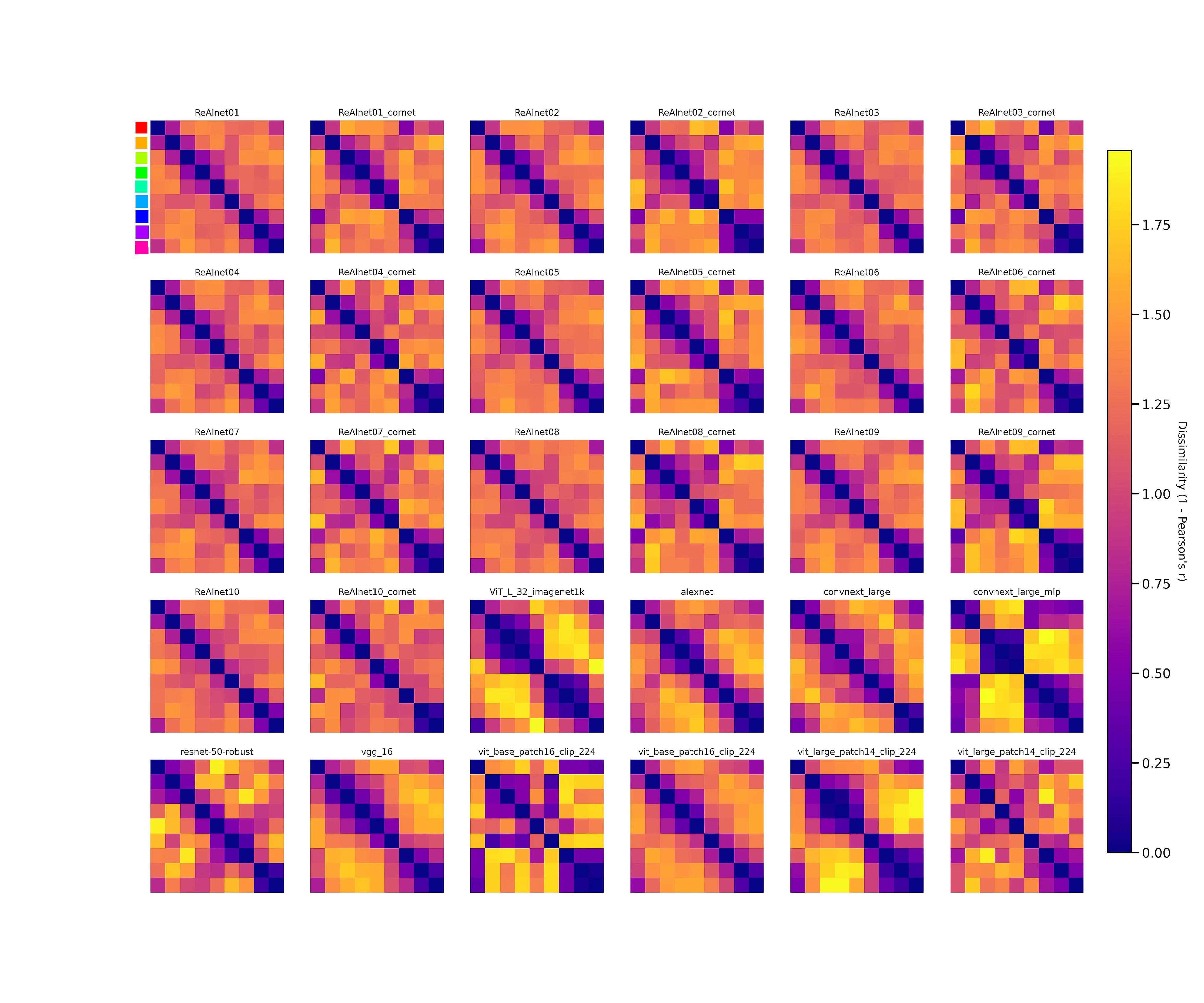}}
\end{figure}



\end{document}